\documentclass[10pt,twocolumn,letterpaper]{article}

\usepackage{wacv}
\usepackage{times}
\usepackage{epsfig}
\usepackage{graphicx}
\usepackage{amsmath}
\usepackage{amssymb}
\usepackage{todonotes}

\usepackage[numbers]{natbib}
\usepackage{multirow}
\usepackage{url}
\usepackage{caption}
\usepackage{subcaption}
\usepackage{xcolor}

\wacvfinalcopy 

\def\wacvPaperID{22} 

\ifwacvfinal\fi
\begin{document}

\title{An Analysis of 1-to-First Matching in Iris Recognition}

\author{Andrey Kuehlkamp \hspace{2cm} Kevin W. Bowyer \\
Department of Computer Science \& Engineering\\
University of Notre Dame\\
Notre Dame, Indiana\\
{\tt\small akuehlka@nd.edu}
}


\maketitle
\ifwacvfinal\thispagestyle{empty}\fi

\begin{abstract}
Iris recognition systems are a mature technology that is widely used throughout the world.
In identification (as opposed to verification) mode, an iris to be recognized is typically matched against all N enrolled irises.
This is the classic "1-to-N search".
In order to improve the speed of large-scale identification, a modified "1-to-First" search has been used in some operational systems. 
A 1-to-First search terminates with the first below-threshold match that is found, whereas a 1-to-N search always finds the best match
across all enrollments.
We know of no previous studies that evaluate how the accuracy of 1-to-First search differs from that of 1-to-N search.
Using a dataset of over 50,000 iris images from 2,800 different irises, we perform experiments to evaluate the relative accuracy of 1-to-First
and 1-to-N search.
We evaluate how the accuracy difference changes with larger numbers of enrolled irises, and with larger ranges of rotational difference
allowed between iris images.
We find that False Match error rate for 1-to-First is higher than for 1-to-N, and the the difference grows with larger number of enrolled irises and with larger range of rotation.\end{abstract}

\section{Introduction}

Iris recognition relies on the capture of images of the colored area of the eye that surrounds the pupil. Each iris contains a complex pattern composed of elements such as freckles, crypts, pits, filaments, furrows, striations and rings. 
Among the advantages of the use of iris recognition, it is possible to highlight: uniqueness of the iris patterns,  robustness against forging and ease of capture and manipulation \citep{Bhattacharyya2009}.


Recent programs like the NEXUS border-crossing in Canada \citep{nexus2014}, the voter registration in Somaliland \citep{bowyer2015iris}, and the Aadhaar program in India \citep{uidai2012}, are examples of large scale application of iris recognition. 
Applications like these point to the need for systems that can operate on a very large number of subjects \citep{Daugman2014}, while maintaining acceptable accuracy and speed. 
In addition, the growth in the research literature indicates the need to solve several issues in order to develop more complex and larger scale systems \citep{Burge2013}.

Larger-scale systems means larger enrollment databases and consequently, for identification systems, longer search times.
The most intuitive way to search a biometric database is through an exhaustive search: the entire database is scanned, and the best match (i.e. the template with smallest Hamming distance) for the probe is selected. 
This process, known as 1-to-N (1:N) search, has the potential disadvantage of long search times.

The initial proposal and following implementations of iris recognition systems were based on the premise of single-image enrollments. 
This was how the field operated for many years, yet many current commercial systems may work with multiple-image enrollments of an iris, which can increase the size of the database and be another factor leading to longer search times.

In order to perform a faster search through a large number of enrolled templates, some applications use what is informally known as 1-to-first (1:First) search. 
In 1:First, the database scan is ended when the first template below some distance threshold is found. 
The key insight here is that once the search has found a first below-threshold match to an enrollment, the system is committed to a match result rather than a non-match result for the presented probe.
Examining the remainder of the enrolled irises is "only" a matter of looking for an even better match to some other enrollment, when such instances "should be" rare.
And from the perspective of the user of the system, 1:First can speed up the process by avoiding the scan of the entire database, reducing the time on average by one half. 
However, the result is not guaranteed to be the best matching of all the enrollments.
This can be a reasonable result because  the user is "recognized" and granted entry, as they should be, even if they are not recognized with the correct identity.

\begin{table*}[t]
	\caption{Comparison of 1:N and 1:First Search Results}
	\label{tab3}
	\begin{center}
		\begin{tabular}{ c | c }
			\hline
			\textbf{Result in 1:N Search} & \textbf{Possible Result(s) if 1:First is used instead}\\ \hline
			\multirow{3}{*}{True Match}  
			& {\parbox{5in}{\vspace{.2\baselineskip}True Match. Occurs if the correct-matching enrollment is the first below-threshold enrollment encountered.\vspace{.2\baselineskip}}}\\ \cline{2-2}
			& {\parbox{5in}{\vspace{.2\baselineskip}False Match. Occurs if the correct matching enrollment is not the first below-threshold enrollment encountered.\vspace{.2\baselineskip}}}\\ \hline

			\multirow{3}{*}{False Match}  
			& {\parbox{5in}{\vspace{.2\baselineskip}False Match. Occurs if correct-matching enrollment is above threshold, or if it is not the first below-threshold enrollment that is encountered. May or may not be the same False Match found in 1:N.\vspace{.2\baselineskip}}}\\ \cline{2-2}
			& {\parbox{5in}{\vspace{.2\baselineskip}True Match. Occurs if the correct-matching enrollment is below-threshold and is encountered before the better-matching but incorrect enrollment found as the false match in 1:N.\vspace{.2\baselineskip}}}\\ \hline

			True Non-Match
			& {\parbox{5in}{\vspace{.2\baselineskip}True Non-Match. Occurs in 1:N if a not-enrolled person uses the system and there is no below-threshold enrollment; 1:First must produce the same result in this case.\vspace{.2\baselineskip}}}\\ \hline

			False Non-Match
			& {\parbox{5in}{\vspace{.2\baselineskip}False Non-Match. Occurs in 1:N if an enrolled person uses the system and there is no below-threshold enrollment; 1:First must produce the same result in this case.\vspace{.2\baselineskip}}}\\ \hline

		\end{tabular}
	\end{center}
\end{table*}

We have found no published work that analyzes the properties of 1:First search in biometrics.
At the same time, 1:First search has been used in multiple large-scale and high-profile iris recognition applications.
Examples of such applications are the Canadian NEXUS border-crossing \citep{Chumakov1} and the Iris Recognition Immigration System (IRIS) in the United Kingdom \citep{IrisUK2014}. 
In this sense, it is of major importance that the differences in behavior between 1:N and 1:First search are well defined and understood.

It can help in understanding 1:First search to catalog when and how it can produce a different result than 1:N search, as summarized in Table \ref{tab3}.
A probe that generates a false match result in 1:N may or may not generate a false match in 1:First.
If the true-matching enrollment is below threshold and is the first below-threshold enrollment encountered, then 1:First search will generate a true match result.
If the correctly matching enrollment is above threshold, or if it is below threshold but would be encountered after the probe that would generate the false match result in 1:N, then 1:First search will generate the same false match result as 1:N.
If there is some other false-matching enrollment that is below threshold and is encountered before the enrollment that would generate the false match in 1:N, then 1:First search will generate a different false match result than 1:N search.

Similarly, an image that will generate a true match result in 1:N may or may not generate a true match in 1:First. 
If the 1:First search encounters any below-threshold match before encountering the true match, it will result in a false match. 
The inverse situation is also possible: a probe that will generate a true match in 1:First can theoretically not generate one in 1:N. 
This could happen if a false match image has the best below-threshold score and is encountered after any below-threshold image. 

A different case takes place for false non-matches: an image that generates a false non-match in 1:N will also generate a false non-match in 1:First. The reason this happens is that in both cases, no match below the threshold is found, after all the images in the gallery are compared.


The purpose of this work is to explore the differences between 1:N and 1:First search, particularly in the context of iris recognition. 
The objective is to assess the impact that using 1:First can have on the accuracy of  iris identification, when compared to the more traditional 1:N search, and especially how the difference scales with the size of the enrollment and with the range of rotation allowed in matching.
The results show that 1:First method generates higher error rates in comparison with 1:N. 
Additionally, the error rates tend to grow as the size of the gallery and/or the number of rotational shifts are increased.

\section{Background}

Iris recognition, as well as other biometric modalities in general, offers two types of identity management functionality \citep{Jain2011}:

\textbf{Verification} is the term employed when a user presents him or herself to the recognition system and claims to be a certain person, and the task of the system is to determine if the claim is true. 
In this case, the biometric template from the user is compared to a single template in the database (\textit{one-to-one matching} \citep{Biometrics2006}). 
If the distance between the two templates meets a determined threshold, the claim is considered "genuine" and the user is accepted. 
If the distance is above such threshold, the user is considered an "impostor" and the claim is rejected.

\textbf{Identification} is the other type of functionality provided by biometric systems. 
This term is used when the user presents him or herself to the recognition system, but does not explicitly claim an identity. 
The system then has to compare the user's biometric template with the templates of potentially all the persons enrolled in the database (\textit{one-to-many matching}\citep{Biometrics2006}). 
The result of this process will be: a) the system "accepts" the user, and assumes that his or her identity is the person with the smallest below-threshold match out of all enrollments in the database; or b) the system "rejects" the user, indicating that the user is not enrolled in the database.

Within identification it is possible to distinguish \emph{closed-set} and \emph{open-set} identification tasks. 
With a closed-set, the user is known to be enrolled in the database, and the system is responsible to determine his or her identity. 
On the other hand, when doing open-set identification, the system must, before trying to identify a user, determine if he or she is enrolled in the database \citep{Biometrics2006}.

This work is concerned with one-to-many matching as used in an identification system, and particularly, with exploring the difference between two possible implementations of one-to-many.

The \emph{matching} procedure is a core part of every biometric identification or verification system. 
In this procedure, the system compares the biometric sample acquired from the user against previously stored templates and scores the level of similarity between them. 
According to a predetermined threshold, the system then makes a decision about the user: either it is a \emph{match} or a \emph{non-match}. 
Declaring a match means to assume that the system accepts both biometric samples as being originated by the same human source \citep{Biometrics2006}.

A biometric system may produce two types of errors, \emph{false match} and \emph{false non-match}. 
A false match occurs when two samples from different individuals are declared by the system as a match. 
A false non-match is when two samples from the same individual fail to be considered as a match by the system \citep{Jain2011}. 

Not every captured image of an iris has the same head tilt, camera angle and rotational torsion of the eye, which can cause it to be misinterpreted as a non-match.
Iris matchers usually offer some tolerance to the iris rotation, in the form of ”best-of-M” comparisons: A comparison between a pair of iris codes is performed several times, over a range of relative rotations, and the best match is chosen to be the distance score for that pair \citep{Daugman2006}. 
As an example, the NEXUS border-crossing program considers a range of 14 rotation values in the initial scan of the enrollment database, but the range is widened to an additional 28 rotation values if no match was found on the initial scan \citep{Chumakov1}.

In practice, the application of biometric identification may encounter restrictions when implementing one-to-many matching. If the enrollment list is large, it may be slow to sweep it entirely every time a user is presented to the system.  
The traditional approach for the implementation of one-to-many identification is the exhaustive 1:N, and is probably the only form of one-to-many matching to receive attention in the research literature.

To speed up the search process in a one-to-many identification, a common approach is to perform a search known as \emph{1-to-first}, in which the system sweeps the enrollment database until it finds the first template for which the distance score is within a defined threshold, and declares a match \citep{Ortiz2015}. 
This approach yields a lower number of comparisons on average, compared with the 1:N method. 
On the other hand, this technique may lead to a higher error rate, since when a match pair is found the search is stopped, ignoring other potentially better matches.

Other efforts have been made in the sense of improving the search performance in iris databases. 
Rakvic \textit{et al.} \citep{rakvic2009} proposed the parallelization of the algorithms involved in the iris recognition process, including the template matching.  
Their parallelized version, although more efficient than a sequential CPU to perform a single match, still has its overall performance directly associated to the size of the database.
In another attempt to address the issue, Hao \textit{et al}. \citep{hao2008} propose an approach based on Nearest Neighbor search, to reduce the search range and thus improve the performance. 

For the sake of clarity, this work will refer to \emph{one-to-many} as the general identification procedure, while considering \emph{1:N} and \emph{1:First} as two different types of one-to-many matching.

\section{Experimental Setup}
\label{Setup}

This work performs an empirical investigation to explore the difference in the accuracy of systems that use the practical approach of \emph{1:First} search, against the more traditional \emph{1:N}.
More specifically, the investigation tries to assess how 1:First scales when applied to a range of different gallery sizes and distance thresholds.
To do so, an environment for \emph{closed-set} identification was set up using an available iris image dataset.

A total of 57,232 iris images were captured with an LG-4000 sensor. An enrollment gallery was created with the earliest image of each iris. 
Then the remaining images of these subjects were used as a probe set to match against the gallery. 
Thus, it is assumed that each iris is enrolled with just one image, and that all probes would correspond to an enrolled identity (a " closed set" scenario).
The ordering of the images in each gallery was defined randomly, and the same order was used in all experiments.

The process was then repeated to create different-sized galleries and probe sets, as shown in Table \ref{tab1}. 
The number of images in each probe set is not uniform, because it tries to maximize the number of probes, but depends on the number of available images for each subject. 
The difference between left and right probe set size is however no larger than 0.15\%.

\begin{table}
	\caption{Galleries and Probe Sets Sizes}
	\label{tab1}
	\begin{center}
		\begin{tabular}{ c | c | c | c }
			\hline
			
			\multicolumn{2}{c|}{\textbf{Left Eye}}	& \multicolumn{2}{c}{\textbf{Right Eye}} \\ \hline
			Gallery & Probe Set & Gallery & Probe Set \\ \hline
			100	& 7,745	& 100	&  7,740	\\ \hline
			200	& 12,529	& 200	& 12,515	\\ \hline
			400	& 18,644	& 400	& 18,624	\\ \hline
			600	& 22,395	& 600	& 22,364	\\ \hline
			800	& 24,555	& 800	& 24,521	\\ \hline
			1,000	& 25,582	& 1,000	& 25,547	\\ \hline
			1,200	& 26,078	& 1,200	& 26,040	\\ \hline
			1,400	& 26,478	& 1,400	& 26,437	\\ \hline
		\end{tabular}
	\end{center}
\end{table}

The iris matcher used to perform the identification was the \emph{IrisBEE} baseline iris recognition algorithm \citep{liu2005}. Analyzing the IrisBEE log file, it is possible to interpret the results in terms of either a 1:N or a 1:First identification.

Since this is a closed-set identification, all of the probes should have a match in the gallery. For the same reason, the number of true non-matches is not considered (An open-set scenario could be a topic for future research).

\begin{figure}[h]
	\centering
	\includegraphics[width=\linewidth]{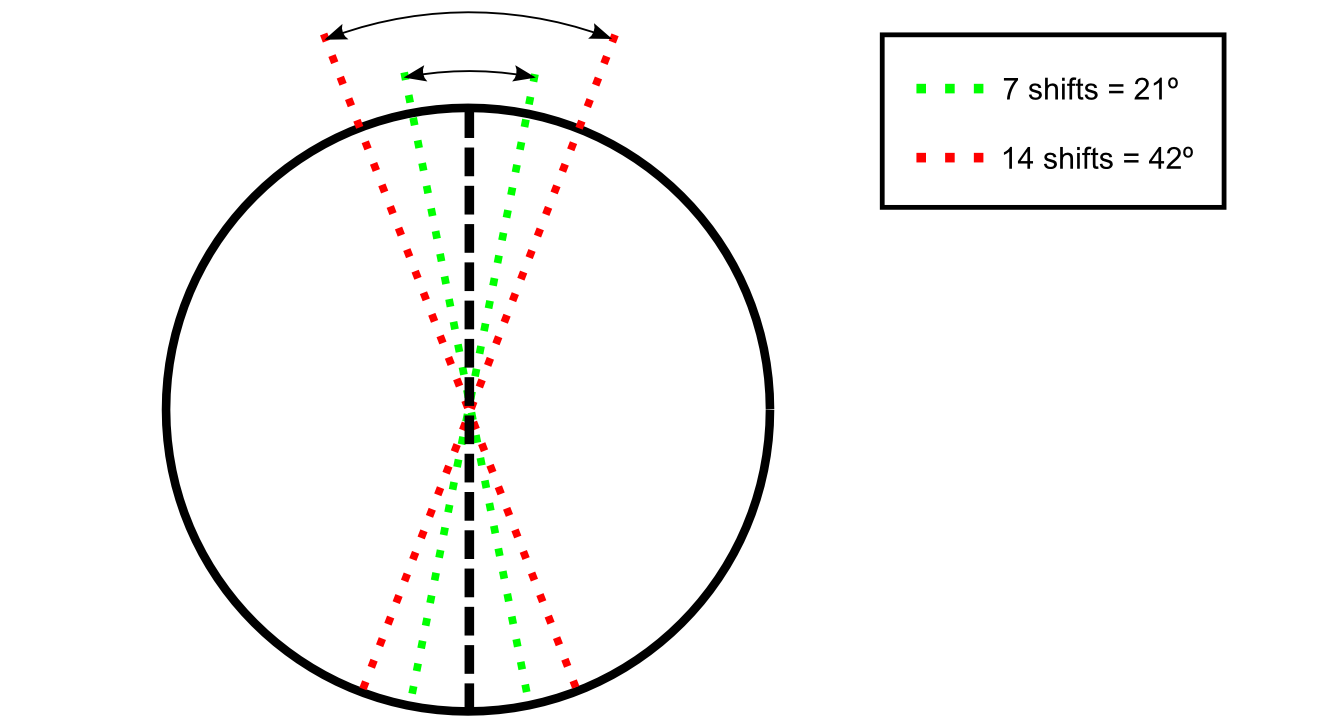} 
	\caption{Iris rotation tolerance limits for the experiments.}
	\label{fig5}
\end{figure}

In order to account for the rotational tolerance, the experiments allowed up to $\pm14$ rotation shifts. 
This number was chosen because it is similar to that used for matching in the NEXUS program. 
In the IrisBEE implementation, each rotation step corresponds to 1.5$^{\circ}$, meaning that rotations of up to 42$^{\circ}$ can be allowed in the most extreme case. Figure \ref{fig5} illustrates this situation.

For each gallery, the matcher tested each probe set using 1:N and 1:First matching. This procedure was repeated using a range of values for the Hamming Distance threshold, and also for the number of eye rotations allowed during matching. 

\section{Results and Discussion}

This section will present and discuss the results obtained in the described experiment. 
After running the matching, the accuracy was measured in terms of performance scores for each of the possible results described in section \ref{Setup}: True Match Rate (TMR), False Match Rate (FMR) and False Non-Match Rate (FNMR). 

\begin{figure}	
	\centering
	\begin{subfigure}[t]{2in}
		\centering
		\includegraphics[width=2in]{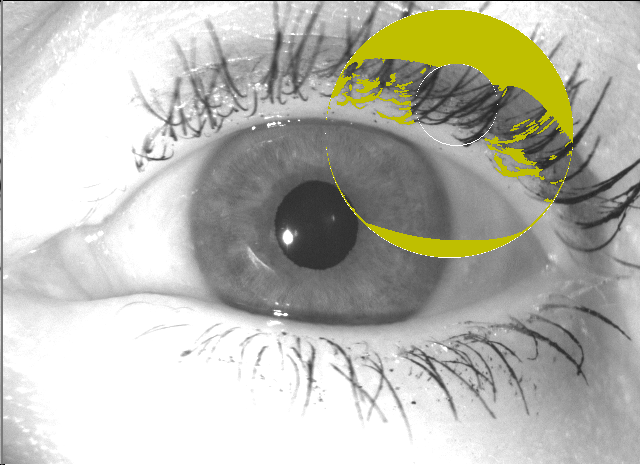}
		\caption{}\label{fig:9a}		
	\end{subfigure}
	\quad
	\begin{subfigure}[t]{2in}
		\centering
		\includegraphics[width=2in]{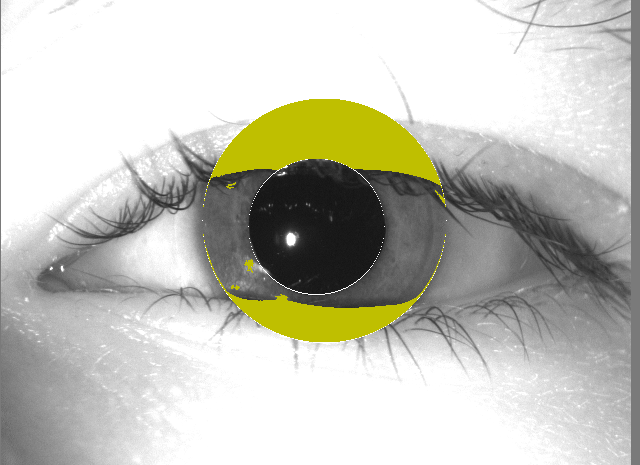}
		\caption{}\label{fig:9b}
	\end{subfigure}
	\caption{Example of false match: despite the evident segmentation error in \ref{fig:9a}, the Hamming Distance to \ref{fig:9b} was 0.298701, a score low enough to be considered a match.}\label{fig:9}
\end{figure}

\begin{figure}	
	\centering
	\begin{subfigure}[t]{2in}
		\centering
		\includegraphics[width=2in]{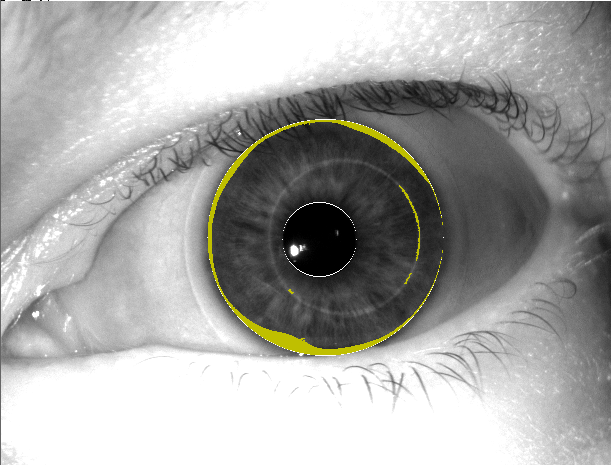}
		\caption{}\label{fig:10a}
	\end{subfigure}
	\quad
	\begin{subfigure}[t]{2in}
		\centering
		\includegraphics[width=2in]{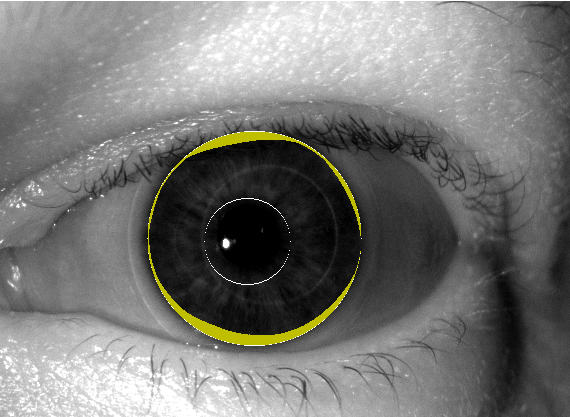}
		\caption{}\label{fig:10b}
	\end{subfigure}
	\caption{Example of false non-match: \ref{fig:10a} and \ref{fig:10b} had a Hamming Distance of 0.423313, and was considered a non-match. The  translucent ring that appears on the subjects' contact lens was partially classified as an occlusion on \ref{fig:10a}, which might have contributed for the high score.}\label{fig:10}
\end{figure}

Figures \ref{fig:9} and \ref{fig:10} show examples of a false match and a false non-match that occurred during the execution of the experiment.

Table \ref{tab2} presents the performance scores for each gallery size, averaged for every threshold between 0.26 and 0.35, in 0.01 increments. From this table, it is easy to perceive that the matching method had little or no effect ($<$0.4\%) in the TMR. 
As for the FNMR, the different matching method had no effect at all. 
This might be explained because most false non matches are the result of factors external to the matching process (e.g. low quality image of the probe or the enrollment).
These external factors can also account for the high FNMR, especially because IrisBEE is not as accurate in the segmentation of the images as other commercial matchers.

\begin{table*}[t]
	\caption{Average Performance Scores (in \%) for Thresholds between 0.26 and 0.35, with increments of 0.01.}
	\label{tab2}
	\begin{center}
		\begin{tabular}{ c|c|c|c|c|c|c }
			\hline
			\multirow{2}{*}{\textbf{Gallery Size}} & \multicolumn{2}{c|}{\textbf{TMR}} & \multicolumn{2}{c|}{\textbf{FMR}} & \multicolumn{2}{c}{\textbf{FNMR}} \\ \cline{2-7}
			& 1:N & 1:First & 1:N & 1:First & 1:N & 1:First \\ \hline
			100 & 86.65 $\pm$ 4.49	&86.65 $\pm$ 4.48	&0.01 $\pm$ 0.03	&0.02 $\pm$ 0.04	&13.33 $\pm$ 4.50	&13.33 $\pm$ 4.50 \\ \hline
			200 & 85.77 $\pm$ 4.55	&85.76 $\pm$ 4.53	&0.02 $\pm$ 0.03	&0.03 $\pm$ 0.05	&14.21 $\pm$ 4.57	&14.21 $\pm$ 4.57 \\ \hline
			400 & 86.21 $\pm$ 4.44	&86.18 $\pm$ 4.40	&0.04 $\pm$ 0.05	&0.07 $\pm$ 0.12	&13.75 $\pm$ 4.48	&13.75 $\pm$ 4.48 \\ \hline
			600 & 86.61 $\pm$ 4.20	&86.65 $\pm$ 4.48	&0.05 $\pm$ 0.08	&0.11 $\pm$ 0.18	&13.34 $\pm$ 4.26	&13.34 $\pm$ 4.25 \\ \hline
			800 & 86.77 $\pm$ 4.13	&86.65 $\pm$ 4.48	&0.06 $\pm$ 0.10	&0.14 $\pm$ 0.24	&13.16 $\pm$ 4.20	&13.17 $\pm$ 4.20 \\ \hline
			1000 & 86.99 $\pm$ 4.05	&86.65 $\pm$ 4.48	&0.08 $\pm$ 0.11	&0.22 $\pm$ 0.38	&12.93 $\pm$ 4.14	&12.93 $\pm$ 4.14 \\ \hline
			1200 & 87.12 $\pm$ 3.99	&86.65 $\pm$ 4.48	&0.10 $\pm$ 0.13	&0.23 $\pm$ 0.36	&12.78 $\pm$ 4.09	&12.78 $\pm$ 4.09 \\ \hline
			1400 & 87.28 $\pm$ 3.94	&86.65 $\pm$ 4.48	&0.12 $\pm$ 0.15	&0.38 $\pm$ 0.58	&12.61 $\pm$ 4.05	&12.61 $\pm$ 4.05 \\ 
			\hline
		\end{tabular}
	\end{center}
\end{table*}

\begin{figure}[h]
	\centering
	\includegraphics[width=\linewidth]{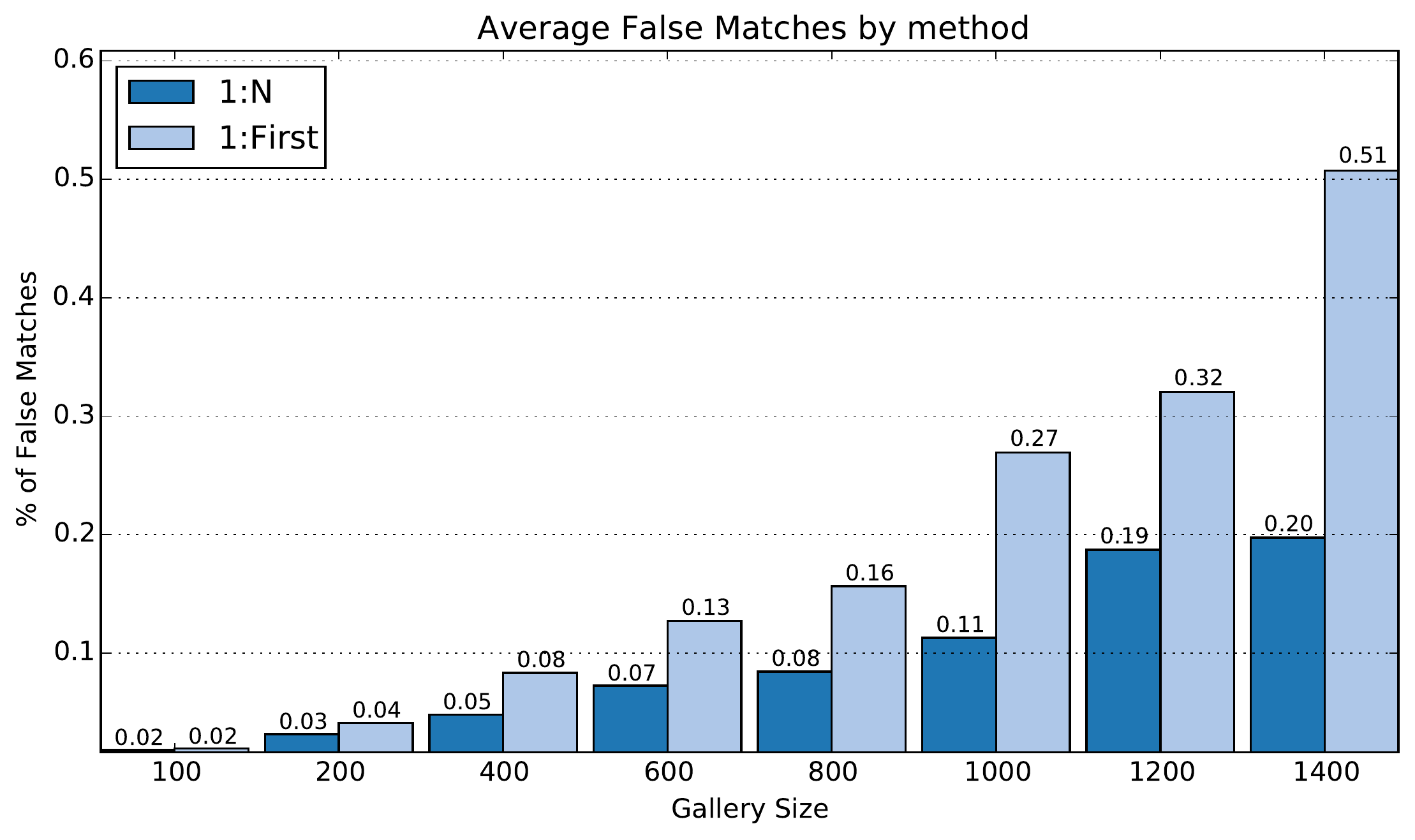} 
	\caption{FMR for 1:N and 1:First as Gallery Size increases.}
	\label{fig1}
\end{figure}

However, the FMR for 1:First identification is up to 2.5 times larger than the FMR for 1:N. 
This fact is illustrated by the graph presented in Figure \ref{fig1}. 
The same Hamming distance thresholds were used for both methods, hence it was expected that the FMR suffered a larger increase for 1:First, when the gallery size was increased. 
Empirically, for the conditions used in this experiment, the rate of increase of errors between 1:N and 1:First grows with the size of the gallery.

Since the FNMR suffered no change, this means that a portion of the comparisons that resulted in True Matches yielded False Matches when using the 1:First method. This fact can be verified observing the correlation between the drop in the TMR and the increase in the FMR.

\begin{figure}[h]
	\centering
		\includegraphics[width=\linewidth]{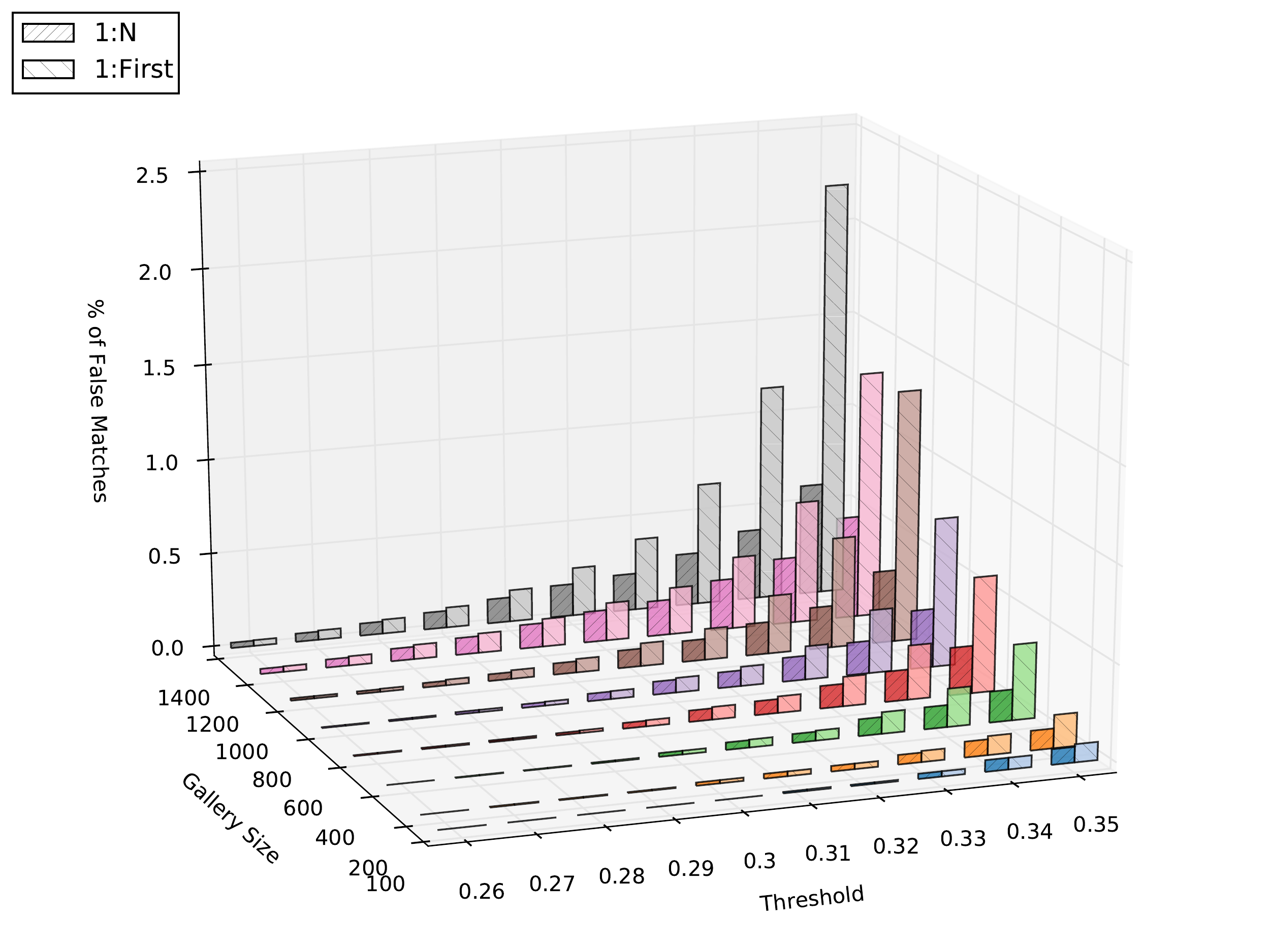}
	\caption{FMR for varying HD and gallery size.}
	\label{fig2}
\end{figure}

To have a better idea of how the 1:First method affects the matching process, experiments with a range of Hamming Distance thresholds and gallery sizes were performed.
Figure \ref{fig2} shows the FMR for each of these experiments, averaged for the left and right eyes. 
In this graph it is easily perceived that the FMR increases both when the threshold is increased and when the gallery size is increased. 
More importantly, it is possible to see that the FMR for 1:First is larger than for 1:N in most cases.

\begin{figure*}
	\centering
		\includegraphics[width=\linewidth]{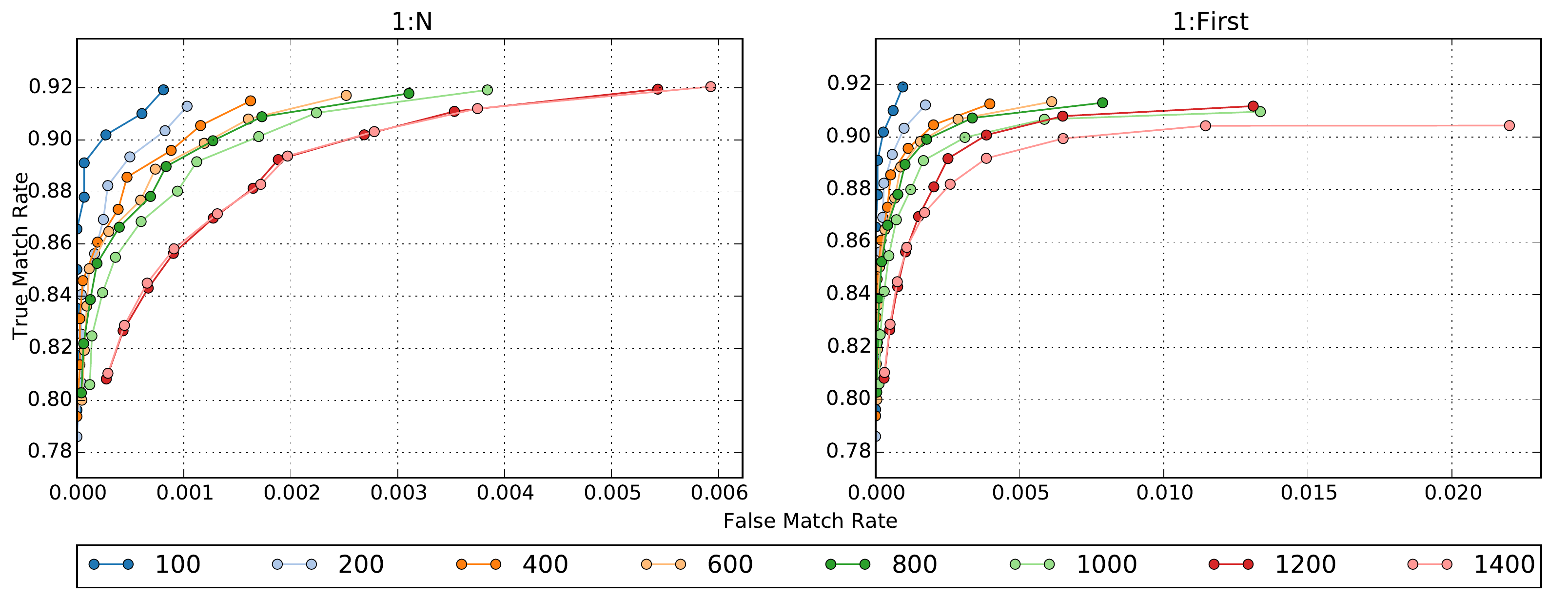} 
	\caption{ROC curve for matching without eye rotation tolerance. The colors represent the gallery size.}
	\label{fig3}
\end{figure*}

Figure \ref{fig3} shows the corresponding ROC charts for both matching methods. 
It is possible also here to notice the interference that the gallery size projects in the performance: in the worst case, 1:N resulted in a TMR of approximately 92\% while maintaining a FMR around 0.6\%. With the same parameters, 1:First produced a TMR a little higher than 90\%, and a FMR of nearly 2.2\%. 
It is also important to notice that in all cases for 1:First, the drop in the TMR when the gallery size is increased reflects directly in an increase of the FMR, but the same does not happen with 1:N matching.

\begin{figure*}
	\centering
	\includegraphics[width=\textwidth]{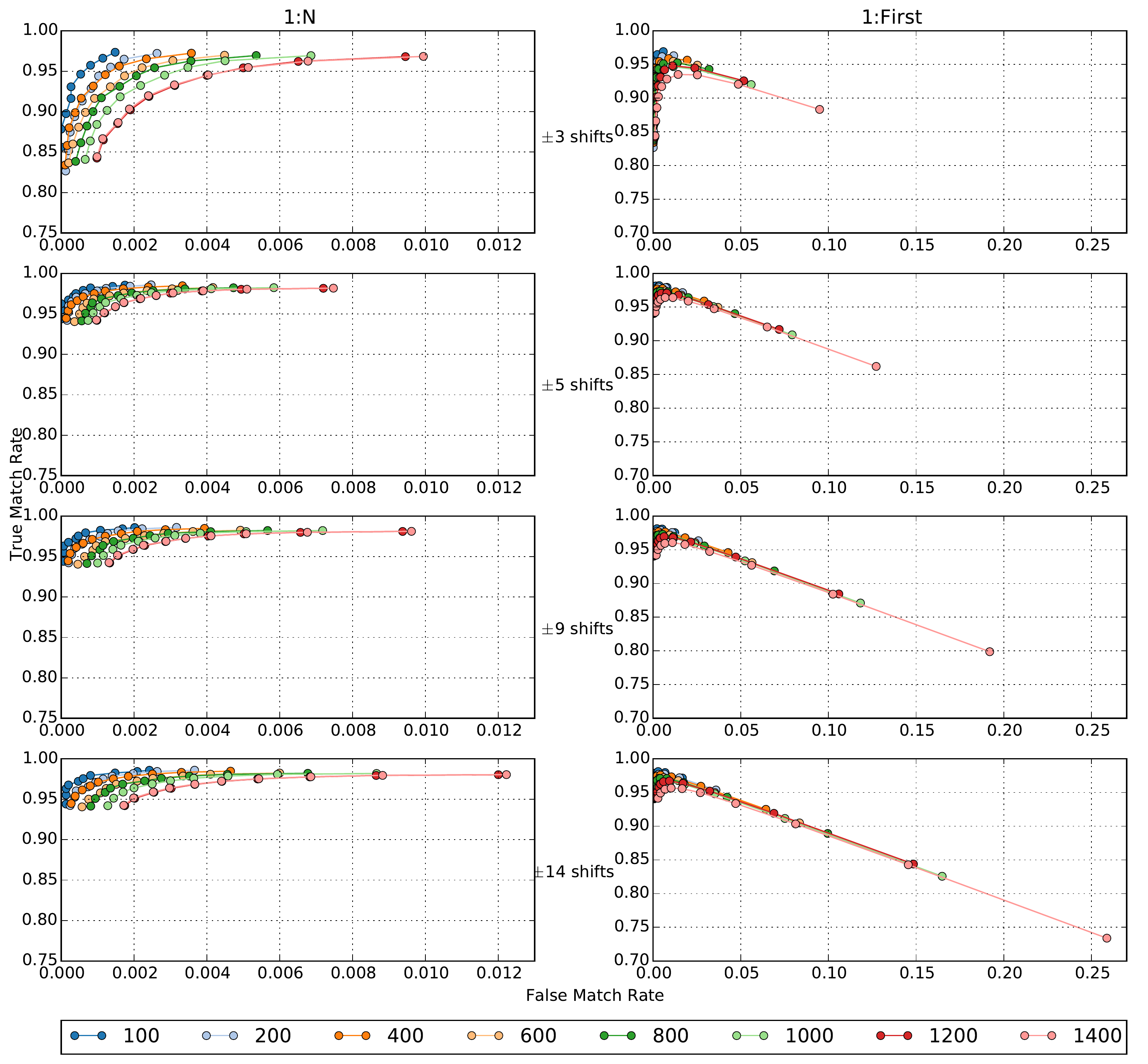}
	\caption{Matching performance with $\pm$3 - 14 rotational shifts. Note that the 1:First ROCs on the right have an unusual aspect. This occurs because the FMR increases and TMR decreases proportionally, with larger threshold values. As the FMR increases with the increase in the threshold, the curve leans to the right; at the same time, the drop in the TMR makes the curve to go down. This effect is particularly accentuated with larger gallery sizes.}
	\label{fig4}
\end{figure*}

The above results were obtained without rotation of the iris codes. 
As explained earlier, relative rotation of the eye between the enrollment and probe is handled by matching the iris codes for a range of relative circular shifts between the codes.
In this case, due to the increase in the number of comparisons, the probability to get a lower Hammming distance score is higher.
In order to get a more realistic assessment, rotation shifts were introduced. 
Figure \ref{fig4} presents the ROC curves obtained for using $\pm$3, 5, 9 and 14 rotational shifts.

The effects of rotational shifts are clearly visible in both cases. 
As the rotation shifts are increased, the TMR lower bound of 1:N rises from around 82\% with $\pm$3 shifts, to close to 94\% with $\pm$9 shifts.
In most of these cases, the TMR is above 98\% for the largest galleries. 
At the same time, the FMR, which is at most 1\% with $\pm$3 shifts, is actually reduced to a little less than 0.8\% when $\pm$5 shifts are used. Additional rotation up to 14 shifts causes the FMR to grow to a little over 1.2\%.

With 1:First, on the other hand, the TMR initially rose to around 96\% for small gallery sizes, but the performance deteriorated for larger galleries, dropping down to below 85\% when using $\pm$9 shifts, and below 75\% with $\pm$14 shifts.
At the same time, the FMR reaches up to nearly 20\% with $\pm$9 shifts and a gallery of 1,400 irises. In its worst case, the FMR reaches approximately 26\% with $\pm$14 shifts. 

\section{Conclusion}

The results point to interesting and impactful issues, especially for large scale applications of iris identification.
Although not much can be inferred by the small changes in the TMR, a closer look at the FMR reveals not only larger values for 1:First, but also an exponential growth curve (Figure \ref{fig1}). 

FNMR was unaffected by the matching method. 
This was expected, because in order for a false non-match to occur, all the enrolled samples must be examined, regardless if using 1:N or 1:First.
On the other hand, there is an inverse relation between the true match and false match rates when using 1:First: as the FMR goes up, the TMR degrades.

Regarding the scaling ability, the results indicate a steep growth in the FMR, as can be seen in Figure \ref{fig2}. This  fact strongly suggests 1:First might not be adequate for large databases, as it tends to generate many more false matches than 1:N.

The analysis of the ROC curves without rotation shifts shown in Figure \ref{fig3} reveals small differences between both methods in the TMR (under 2\%), but with the exception of the two smaller galleries, 1:First is the lowest in all cases.
Accordingly, the False Match Rate of 1:First is higher in nearly all cases, and it tends to get worse with the gallery size growth. In the worst case, the FMR for 1:First is more than 3 times higher than 1:N.

When rotation shifts are introduced (Figure \ref{fig4}), the problems with 1:First become more evident. 
In 1:N matching, what happens as the number of allowed rotation shifts is increased, is that the TMR is increased up to a little more than 94\% in the worst scenario, in comparison with the results shown in Figure \ref{fig3}. 
At the same time, the FMR goes up to 1.2\% in the worst case. 
The rotational shifting has improved the TMR without increasing the FMR. 

In comparison, when the same number of rotations is used in 1:First, the TMR is increased for smaller galleries (less than 800 images), but at the same time the FMR begins to increase proportionally. 
Ultimately, the FMR gain ends up forcing down the TMR, and yielding worse results.
For the larger galleries, 1:First had TMR's below 90\%, with a FMR up to 25\%.

The experiment showed that the behavior of 1:First matching is not as similar to 1:N as might be expected, and unfortunately for the worse.
The low performance of 1:First in the false match scores when compared to 1:N would, by itself, argue against the method. 
But it is with larger galleries and the addition of rotational shifts that 1:First performance was really disturbed, raising the question if the reduction in search time could outweigh the loss in accuracy.

In conclusion, results indicate that while the FNM errors are the same for both 1:N and 1:First, FM errors are quite different. 
Furthermore, the FM errors are generically larger in 1:First than in 1:N, and the problem is worsened by 1) larger gallery sizes, and 2) more shifts to handle a wider range of eye rotation.
Although the IrisBEE matcher is not as accurate as commercial level matchers, one can still expect the same basic trends would be shown in a commercial matcher, but with higher overall levels of accuracy.
Future work could include experiments with an "industrial strength" matcher, with results averaged over multiple random orderings of the enrollment databases, and consideration of an "open set" matching scenario.

\section{Acknowledgements}

This research was partially supported by 
the Brazilian Ministry of Education -- CAPES through process BEX 12976/13-0.

{\small
\bibliographystyle{ieee}
\bibliography{egbib}
}

\end{document}